\documentclass[a4paper,reqno]{amsart}

\usepackage[foot]{amsaddr}
\usepackage[english]{babel}
\usepackage[utf8]{inputenc}
\usepackage{csquotes}
\usepackage{amsmath,amssymb,amsfonts,amsthm}
\usepackage{bm}
\usepackage{enumerate}
\usepackage{booktabs}
\usepackage{graphicx}
\usepackage{tikz}
\tikzset{>=stealth}
\usetikzlibrary{shapes,arrows,shadows,snakes}
\usepackage{longtable}
\usepackage{subcaption}
\usepackage{caption}
\usepackage{setspace}

\usepackage[colorlinks=true,citecolor=blue]{hyperref}
\usepackage[
maxbibnames=99,
maxcitenames=2,
style=numeric,
backend=bibtex,
sorting=nyt,
sortcites,
giveninits=false,
doi=false,
url=true,
natbib=true]{biblatex}

\usepackage{algorithm}
\usepackage[noend]{algpseudocode}
\makeatletter
\def\BState{\State\hskip-\ALG@thistlm}
\makeatother

\theoremstyle{plain}

\theoremstyle{remark}

\theoremstyle{definition}

\title{Model selection in reconciling hierarchical time series}
\author[Mahdi. Abolghasemi]{M. Abolghasemi$^{1}$}
\author[Rob. J. Hyndman]{R. J. Hyndman$^2$}
\author[Evangelos. Spiliotis]{E.Spiliotis$^3$}
\author[Christoph. Bergmeir]{C.Bergmeir$^4$}
\address{$^1$Department of Data Science and AI, Monash University, VIC, Australia}
\address{$^2$Department of Econometrics and Business Statistics, Monash University, VIC, Australia}
\address{$^3$Forecasting and Strategy Unit, National Technical University of Athens, Greece}
\address{$^4$Department of Data Science and AI, , Monash University, VIC, Australia}
\email[M.~Abolghasemi]{mahdi.abolghasemi@monash.edu}
\email[R.J.~Hyndman]{rob.hyndman@monash.edu}
\email[E.S~Spiliotis]{spiliotis@fus.gr}
\email[C.B.~Bergmeir]{christoph.bergmeir@monash.edu}

\date{}

\begin{document}
	
\begin{abstract} 
	Model selection has been proven an effective strategy for improving accuracy in time series forecasting applications. However, when dealing with hierarchical time series, apart from selecting the most appropriate forecasting model, forecasters have also to select a suitable method for reconciling the base forecasts produced for each series to make sure they are coherent. Although some hierarchical forecasting methods like minimum trace are strongly supported both theoretically and empirically for reconciling the base forecasts, there are still circumstances under which they might not produce the most accurate results, being outperformed by other methods. In this paper we propose an approach for dynamically selecting the most appropriate hierarchical forecasting method and succeeding better forecasting accuracy along with coherence. The approach, to be called conditional hierarchical forecasting, is based on Machine Learning classification methods and uses time series features as leading indicators for performing the selection for each hierarchy examined considering a variety of alternatives. Our results suggest that conditional hierarchical forecasting leads to significantly more accurate forecasts than standard approaches, especially at lower hierarchical levels. 
\end{abstract}
	
	
\maketitle

\section{Introduction and background} \label{sec:literaturehierarchy}

Forecasting is essential for supporting decision making, especially in applications that involve a lot of uncertainty. For instance, accurately forecasting the future demand of stock keeping units (SKUs) can significantly improve supply chain management \citep{GHOBBAR20032097}, reduce inventory costs \citep{SYNTETOS2010134}, and increase service levels \citep{pooya2019exact}, particularly under the presence of promotions \citep{ALI200912340, abolghasemi2020demand2}. In order to obtain more accurate forecasts, forecasters typically try to identify the most appropriate forecasting model for each series from a variety of alternatives. This task, although it can provide significant improvements under perfect foresight \citep{fildes2001537}, is difficult to effectively perform in practice due to model, parameter, and data uncertainty \citep{PETROPOULOS2018545}. Thus, many strategies have been proposed in the literature to effectively perform forecasting model selection \citep{FILDES20151692}, most of which are based on the in-sample and out-of-sample accuracy of the forecasting models \citep{TASHMAN2000437}, their complexity \citep{HYNDMAN2002439}, and the features that time series display \citep{Petropoulos2014a, MONTEROMANSO202086}.

However, in business forecasting applications, data is typically grouped based on its context and characteristics, thus structuring cross-sectional hierarchies. For example, although the demand of an SKU can be reported at a store level, it can be also aggregated (summed) at a regional or national level. Similarly, demand can be aggregated for various SKUs of the same type (e.g., dairy products) or category (e.g., foods). As a result, hierarchical time series introduce additional complexity to the whole forecasting process since, apart from selecting the most appropriate forecasting model for each series, forecasters have also to account for coherence, i.e., make sure that the forecasts produced at the lower hierarchical levels will sum up to those produced at the higher ones \citep{Athanasopoulos2020-cx}. In fact, coherence is a prerequisite in hierarchical forecasting (HF) applications as it ensures that different decisions made across different hierarchical levels will be aligned. 

Naturally, the demand recorded at lower hierarchical levels will always add up to the observed demand at higher levels. However, this is rarely the case for forecasts which are usually produced for each series separately and are therefore incoherent. To achieve coherence, various HF methods can be used for reconciling the individual, base forecasts \citep{Spiliotis2019-hg}. The most basic HF method is probably the bottom-up (BU), according to which base forecasts are produced just for the series at the lowest level of the hierarchy, being then aggregated to provide forecasts for the series at the higher levels \citep{dangerfield1992top}. Top-down (TD) is another option which suggests forecasting just the series at the highest level of the hierarchy and then using proportions to disaggregate these forecasts and predict the series at the lower levels \citep{gross1990disaggregation, athanasopoulos2009hierarchical}. Middle-out (MO) mixes the above-mentioned methods, producing base forecasts for a middle level of the hierarchy and then aggregating or disaggregating them to forecast the higher and lower levels, respectively. Finally, a variety of HF methods that combine (COM) the forecasts produced at all hierarchical levels have been proposed in the literature succeeding more often than not, along with coherence, better forecasting performance \citep{hyndman2011optimal, wickramasuriya2019optimal, Jeon2019-xo}.

From the HF methods found in the literature, a COM method, called minimum trace (MinT; \cite{wickramasuriya2019optimal}), has been distinguished due to the strong theory supporting it and the results of many empirical studies highlighting its merits over other alternatives. However, there are still circumstances under which MinT might fail to provide the most accurate forecasts. For instance, since MinT is based on the estimation of the one-step-ahead error covariance matrix, the method might be proven inappropriate when the in-sample errors of the baseline forecasting models do not represent post-sample accuracy, the assumption that the multi-step forecast error covariance is proportional to the one-step forecast error covariance is unrealistic, or the required estimations are computationally too hard to make. Moreover, since MinT treats all levels equally, it cannot be optimized with respect to certain hierarchical levels of interest. Finally, given that medians are not additive, there is no reason to expect that MinT will always improve  forecasting accuracy, at least when measures that are based on averages of forecast errors are used for evaluating forecasting performance.

In such cases, simpler HF methods like the BU and the TD can be proven useful. However, there might still be inadequate evidence about which of the two methods to use \citep{hyndman2011optimal}. For example, the BU method is typically regarded more suitable for short-term forecasts and for hierarchies which bottom series are not highly correlated but less robust to randomness. On the other hand, the TD method is usually regarded more appropriate for long-term forecasts but less accurate for predicting the series at the lower aggregation levels due to information loss. It seems that no reconciliation method can fit all kinds of HF problems and that, similarly to forecasting model selection, the appropriateness of the different HF methods depends on various factors, including the particularities of the time series \citep{nenova2016determining} and the structure of the hierarchy \citep{Fliedner1999, Fliedner2001, gross1990disaggregation, abolghasemi2019machine}, among others. 

The above findings reconfirm George Box's quote that ``all models are wrong, but some are useful'' \cite{quote}, and highlight the potential benefits of conditional HF, i.e., the improvements in terms of forecasting accuracy that could be possibly achieved if forecasters were able to select the most appropriate HF method according to the characteristics of the series that form a hierarchy. In this paper we propose an approach for performing such a conditional selection using time series features as leading indicators \citep{KANG2017345, SPILIOTIS202037} and Machine Learning (ML) methods for conducting the classification. Essentially, we suggest that the forecasting accuracy of the different HF methods found in the literature is closely related with the characteristics of the individual series and that, based on these relationships, ``horses for courses'' can be effectively identified  \citep{Petropoulos2014a}.

We benchmark the accuracy of the proposed approach against various HF methods, both standard and state-of-the-art, using a large dataset coming from the retail industry. Our results suggest that conditional HF leads to superior forecasts which are significantly better than those of the individual HF methods examined. Thus, we conclude that selection should not be limited to forecasting models, as done till now, but be expanded to HF methods as well.

The remainder of the paper is organised as follows. Section~\ref{sec:HFmodels} describes the most popular HF methods found in the literature and Section~\ref{sec:conditionalhier} introduces conditional HF. Section~\ref{sec:setupdata} presents the dataset used for the empirical evaluation of the proposed approach and describes the experimental set-up. Section~\ref{sec:hierarchicalresults} presents the results of the experiment and discusses our findings. Finally, Section~\ref{sec:hierarchicalconclusion} concludes the paper.

\section{Hierarchical forecasting methods} \label{sec:HFmodels}

In this section, we discuss the TD, BU, and COM as three well-established HF methods that are widely used in the literature and in practice for reconciling hierarchical base forecasts. These methods are also the ones considered in this study, both as alternatives of the conditional HF approach to be described in the next section and as benchmarks. For a more detailed discussion on the existing HF methods, their advantages, and drawbacks, please refer to the study of \cite{Athanasopoulos2020-cx}. 

Before proceeding, we introduce the following notations and parameters that will facilitate the discussion of the three methods:

\begin{align*}
&m                     : \text{total number of series in the hierarchy} \\
&m_i                   :\text{ total number of the series for level $i$}; \\
&k                     : \text{total number of the levels in hierarchy}; \\
&n                     : \text{number of the observations in each series}; \\
&Y_{x,t}               : \text{the $t^{th}$ observation of series $Y_x$}; \\
&\hat{Y}_{x,n} (h)    : \text{$h$-step-ahead independent base forecast of series $Y_x$ based on $n$ observations}; \\
&\bm{Y}_{i,t}          : \text{the vector of all observations at level $i$}; \\
&\hat{\bm{Y}}_{i,t}(h) : \text{$h$-step-ahead forecast at level $i$}; \\
&\bm{Y}_t              : \text{a column vector including all observations}; \\
&\hat{\bm{Y}}_n (h)    : \text{$h$-step-ahead independent base forecast of all series based on $n$ observations}; \\
&\tilde{\bm{Y}}_n (h)  : \text{the final reconciled forecasts of all series}
\end{align*}

\addtocounter{table}{-1}

We can express a hierarchical time series as $\bm{Y}_t = \bm{S} \bm{Y}_{k,t}$, where $\bm{S}$ is a summing matrix of order $m \times m_k$.  For example, we can express the three-level hierarchical time series shown in Figure~\ref{htsstructure:sample} as:

\begin{figure}
	\centerline{\includegraphics[scale=0.55]{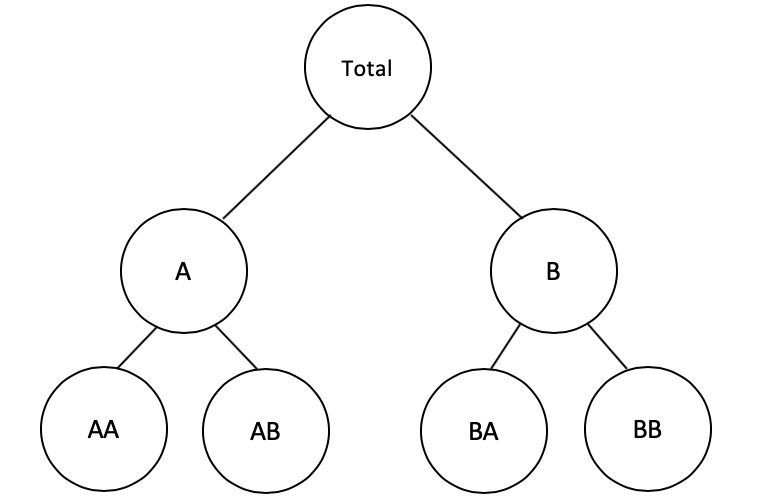}}
	\caption{A three-level hierarchical structure}
	\label{htsstructure:sample}
\end{figure}


\begin{equation*}
\left[
\begin{array}{c}
Y_t \\
Y_{A,t} \\
Y_{B,t}\\
Y_{AA,t} \\
Y_{AB,t}\\
Y_{BA,t} \\
Y_{BB,t} \\
\end{array}
\right] =
\left[
\begin{array}{@{}*{4}{c}@{}}
1 & 1 & 1        & 1 \\
1 & 1 & 0        & 0 \\
0 & 0 & 1        & 1 \\
&   & \bm{I}_4 & \\
\end{array}
\right]
\times
\left[
\begin{array}{c}
Y_{AA,t} \\
Y_{AB,t} \\
Y_{BA,t} \\
Y_{BB,t} \\
\end{array}
\right]
\end{equation*}

Accordingly, we can express various hierarchical structures with a unified format as $\tilde{\bm{Y}}_n (h)= \bm{S} \bm{G} \hat{\bm{Y}}_n(h)$, where $\bm{G}$ is a matrix of order $m \times m_k$ which elements depend on the type of the reconciliation method used, in our case the BU, TD, and COM methods \cite{fpp2}.

\subsection{Bottom-up}\label{buhier}
BU is the simplest HF method according to which we forecast the series at the bottom level of the hierarchy and then aggregate these forecasts to obtain forecasts at higher levels. In this case, the matrix $\bm{G}$ can be constructed as $\bm{G}= [\bm{0}_{m_k \times (m - m_k)}| \bm{I}_{m_k}]'$, where $\bm{0}_{i \times j}$ is a $i \times j $ null matrix. 

\subsection{Top-down}\label{tdhier}
In the TD method, base forecasts are produced at the top level of the hierarchy and then disaggregated to the lower levels with appropriate factors. While there are various ways for computing such factors and disaggregating the top level forecasts, we consider the proportions of the historical averages since it is a widely used  alternative that provides reasonable results \citep{athanasopoulos2009hierarchical}. These proportions are computed as follows 

\begin{align}
\label{td2}
p_j = \frac{\sum_{t=1}^{n} Y_{j,t}}{\sum_{t=1}^{n} Y_t}, ~~~~~~~~  j=1, \dots, m_k
\end{align}

\noindent where $p_j$ represents the average of the historical value of the bottom level series ${Y_{j,t}}$ relative to the average value of the total aggregate ${Y_t}$. We can then construct the vector $\bm{g} =[p_1, p_2, p_3, \dots, p_{m_k}]$ and matrix $\bm{G}= [\bm{g} \mid \bm{0}_{m_k \times (m - 1)}]'$. 

\subsection{Optimal combination}\label{optimalhier}

The COM method produces base forecasts for all series across all hierarchical levels and then combines them with a linear model to obtain the reconciled forecasts. Suppose $$\tilde{\bm{Y}}_n(h) = \bm{S} \bm{G} \hat{\bm{Y}_n}(h) $$ depicts the $h$-step-ahead reconciled forecasts. Then, the covariance matrix of the errors of these forecasts can be given by $$\bm{V}_h=\text{Var}[\bm{y}_{n+h} - \tilde{\bm{Y}}_n (h)]=\bm{S}\bm{G}\bm{W}_h\bm{G}'\bm{S}',$$

\noindent where $\bm{W}_h$ is the variance-covariance matrix of the $h$-step-ahead base forecast errors \citep{wickramasuriya2019optimal, hyndman2016fast}. It can be shown that the matrix $\bm{G}$ that minimizes the trace of $\bm{V}_h$ such that it generates unbiased reconciled forecasts, i.e., $\bm{S}\bm{G}\bm{S}=\bm{S}$, is given by $$\bm{G}= (\bm{S}' \bm{W}^{\dagger}_h\bm{S})^{-1} \bm{S}' \bm{W}_h^{\dagger},$$ where $\bm{W}^\dagger_h$ is the generalized inverse of $\bm{W}_h$.



%

There are a few different ways to estimate $\bm{W}_h$. In this study we consider the shrinkage estimation as it has been empirically shown that it yields the most accurate forecasts in many HF applications \cite{abolghasemi2019machine, spiliotis2020hierarchical, wickramasuriya2019optimal}. Using the shrinkage method, this matrix can be estimated by $\bm{W}_h=k_h\left(\lambda_D \hat{\bm{W}}_{1,D} + (1-\lambda_D)\hat{\bm{W}}_1\right)$.  The diagonal target of the shrinkage estimator is $\hat{\bm{W}}_{1,D} = \text{diag}(\hat{\bm{W}}_1)$ and the shrinkage parameter is given by $$\lambda_D = \frac{\sum_{i \neq j} \text{Var}(\hat{r}_{ij})}{\sum_{i \neq j} \hat{r}^2_{ij}},$$ 	where $\hat{r}_{ij}$ is the $(i,j)^{th}$ element of the one-step-ahead in-sample correlation matrix \cite{schafer2005shrinkage}.  

The COM method was implemented using the \texttt{MinT} function of the \textit{hts} package for R \cite{htspackage}.

\section{Conditional hierarchical forecasting} \label{sec:conditionalhier}

Time series often depict different patterns, such as seasonality, randomness, noise, and auto-correlation \citep{KANG2017345}. As a result, there is no model that can consistently forecast all types of series more accurately than other models, even relatively simple ones \citep{Petropoulos2014a, FILDES20151692}. Similarly, although some models may perform better on a time series dataset of particular characteristics, there is no guarantee that this will always be the case \citep{SPILIOTIS202037}. For example, despite that exponential smoothing \citep{Gardner1985} typically produces accurate forecasts for seasonal series, it might be outperformed by ML methods when a large number of observations is available \citep{SMYL202075}. Thus, selecting the most appropriate forecasting model for each series becomes a promising, yet challenging task for improving overall forecasting accuracy \citep{MONTEROMANSO202086}. 

Model selection has been extensively studied in the forecasting literature. Although there is no unique way to determine the most appropriate forecasting model for each series, empirical studies have provided effective strategies for performing this task \citep{FILDES20151692}. From these strategies, the approaches that build on time series features are among the most promising ones given that the latter can effectively represent the behaviour of the series in an abstract form and match it with the relative performance of various forecasting models \cite{reid1972comparison, meade2000evidence, wang2009rule, Petropoulos2014a, KANG2017345, abolghasemi2020demand}. 

Expert systems and rule-based forecasting were two of the early approaches to be suggested for forecasting model selection \cite{collopy1992rule, mahajan1988new}. \citet{collopy1992rule} considered domain knowledge along with 18 time series features and proposed a framework that consisted of 99 rules to select the most appropriate forecasting model from 4 alternatives. In another study, \citet{adya2000application} considered 6 features and 64 rules to select the most accurate forecasting model from 3 alternatives. Similarly, \citet{adya2001automatic} proposed an approach to automatically extract time series features and choose the best forecasting model. \citet{Petropoulos2014a} measured the impact of 7 time series features plus the length of the forecasting horizon on the accuracy of 14 popular forecasting models, while \citep{KANG2017345} and \citep{SPILIOTIS202037} linked the performance of standard time series forecasting models with that of various indicative features using data from well-known forecasting competitions. More recently, \citet{MONTEROMANSO202086} used 42 time series features to determine the weights for optimally combining 9 different forecasting models, winning the second place in the M4 forecasting competition \citep{MAKRIDAKIS202054}.

Inspired by the work done in the area of forecasting model selection, we posit that HF methods can be similarly selected using time series features and, based on such a selection, improve forecasting accuracy for the case of hierarchical series. In this respect, we proceed by computing various time series features across all hierarchical levels and propose using these features for selecting the HF method that best suites the hierarchy, i.e., produces on average the most accurate forecasts for all the series it consists of \cite{abolghasemi2020demand, Petropoulos2014a}. The selection is done by employing a popular ML classification method. 

\begin{figure}
	\caption{CHF algorithm flowchart}
	\includegraphics[scale=0.4]{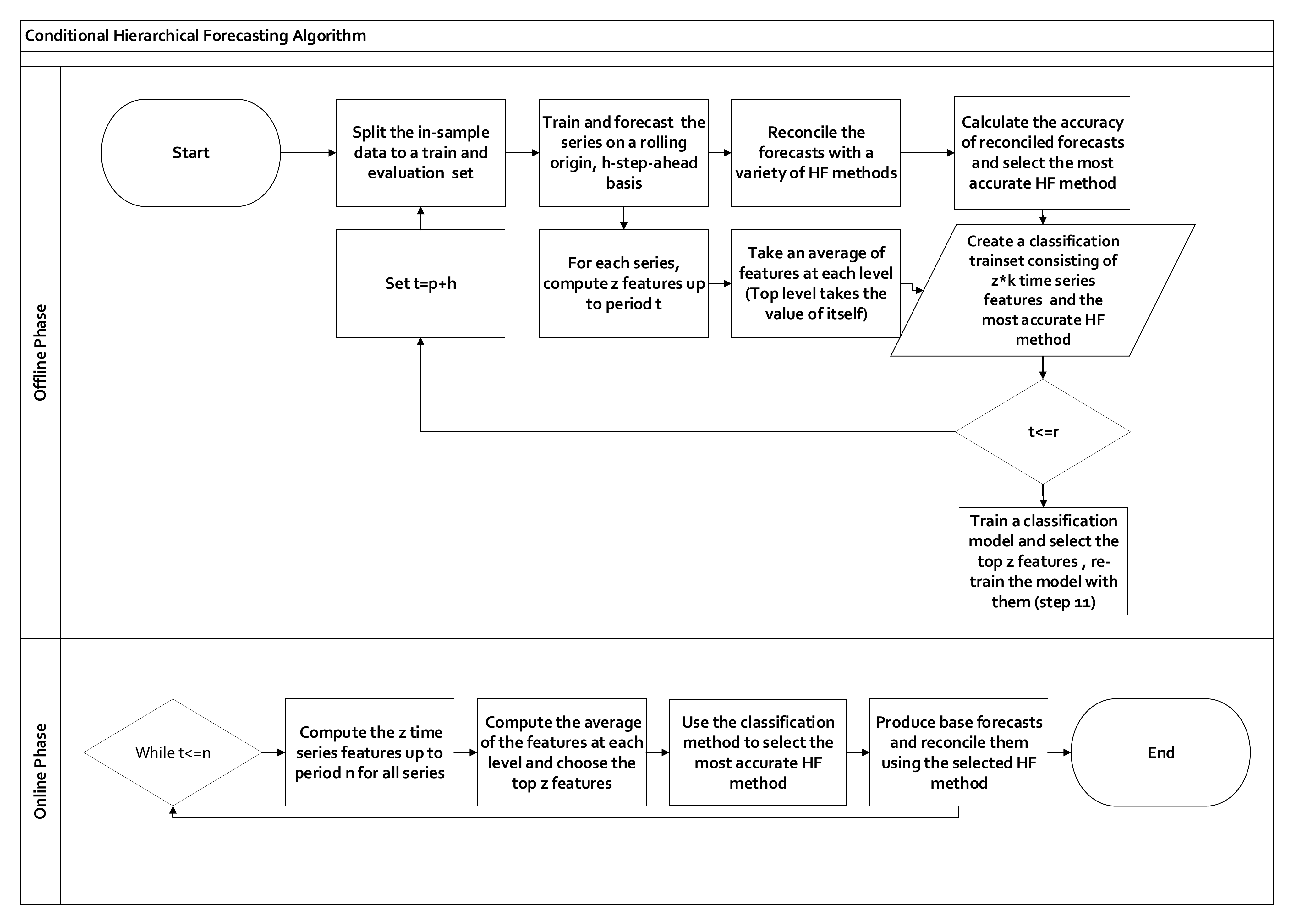}
	\label{algo}
\end{figure}

The proposed approach, to be called conditional HF (CHF), is summarised below and presented in the flowchart of Figure \ref{algo}.

\begin{algorithm}[H]
	\begin{algorithmic}[1]
		\State 	\textbf{Off-line phase}
		\For {$t =p ~to ~r$ with a step of $h$} 
		\State {Create a train and test set by splitting the available in-sample data of size $r$. The train set includes the first $p$ observations and the test set includes the following $h$ observations, $p+1:p+h$, equal in number with the forecasting horizon considered.} 
		\State Fit a forecasting model of choice to the train set and produce $h$-step-ahead forecasts. 
		\State Reconcile the base forecasts produced with different HF methods of choice.
		\State Calculate the accuracy of the $h$-step-ahead forecasts produced by the different HF methods using a measure of choice.
		\State Compute a variety of time series features ($z$ in number) for the $m$ series of the hierarchy.\EndFor
		\State Create a train set for a ML classification method of choice. The train set of the classification method includes the average values of the time series features considered in step 7, computed across all the series of each hierarchical level separately. Thus, a total of $k \times z$ independent numerical variables are provided as input to the classifier. The target variable is the most accurate reconciliation method, as determined in step 6, and is provided to the classifier as a categorical variable.
		\State Train the ML classification method using the train set developed in step 8.
		\State \textbf{On-line phase}
		\For {$t = r+1 ~to ~ n$ with a step of $h$} 
		\State Compute the time series features considered in step 7 for all the $m$ series of the hierarchy up to observation $t$. Then, compute the average value of these features per hierarchical level, as done in step 8. 
		\State Use the classification method trained in step 9 and the input data constructed in step 12 to predict the class of the examined hierarchy, i.e., the most accurate HF method from the ones considered in step 5.
		\State Produce base forecasts for the following $h$ periods, $t+1:t+h$, and reconcile these forecasts using the HF method predicted in step 13.\EndFor
	\end{algorithmic}
	\caption*{\textbf{Algorithm} Conditional Hierarchical Forecasting}
	\label{CHF Algorithm}
\end{algorithm}

Given that CHF builds on time series features and its accuracy is directly connected with the representativeness of the features used, as well as the capacity of the algorithm employed for selecting the most appropriate HF method, it becomes evident that choosing a set of diverse, yet finite features is a prerequisite for enhancing the performance of the proposed classification approach. There are many features that can be used to describe time series patterns. For example, \cite{fulcher2013highly} extracted more than 7,700 features for describing the behaviour of the time series and then summarised them into 22 canonical features, losing just 7\% of accuracy \cite{catch22}. Similarly, \cite{wang2006characteristic} and \cite{KANG2017345} suggested that a relatively small number of features can be used for effectively visualising time series and performing forecasting model selection. Based on the above, we decided to consider 32 features for the CHF algorithm so that the patters of the hierarchical series are effectively captured without exaggerating. These features included \textit{entropy, lumpiness, hurst, acffeatures, pacffeatures, nperiods, seasonal-period, trend, spike, linearity, curvature,e-acf1, e-acf10, seasonal-strength, x-acf1, x-acf10, diff1-acf1, diff1-acf10, diff2-acf1, diff2-acf10, seas-acf1, x-pacf5, diff1x-pacf5, diff2x-pacf5, seas-pacf, linearity and non-linearity, arch-test, luctanal-prop-r1, and unitroot-kpss}, and were computed using the \textit{tsfeatures} package for R \cite{tsfeatures}. For more details about these features, please refer to the studies of \cite{wang2006characteristic} and \cite{KANG2017345}.

Note that CHF is flexible in terms of the method that will be employed for performing the classification. That is, users can choose their classification method of choice for identifying the most accurate HF method and reconciling the base forecasts produced for the examined hierarchy. In the preliminary experiment, we considered three methods, namely eXtreme Gradient Boosting (XGB), Support Vector Machines (SVM), and Random Forests (RF). However, since XGB performed best, we decided to exclude the last two from our analysis. This is aligned with the mainstream knowledge in data science \cite{nielsen2016tree, chatzis2018forecasting, demolli2019wind}. XGB is an implementation of gradient boosted decision trees that is based on ensembling and uses a large number of hyper-parameters for performing the classification. Since XGB results are fairly robust to the chosen values of the hyper-parameters, we set their values manually by considering the literature and users' experience \cite{spiliotis2020hierarchical}. More specifically, we set the learning rate (\texttt{eta}) to 0.01, \texttt{colsample-bytree} to 1, \texttt{min-child weight} to 5 , \texttt{max-depth} to 5, \texttt{sub sample} size to 0.7, and the number of iterations to 1000. We trained XGB by setting the objective to \textit{multi:softprob} and evaluated its performance with \textit{mlogloss}.

Note that CHF is also flexible in terms of the forecasting models that will be used for producing the base forecasts and the HF methods that will be considered for their reconciliation. For the latter case, in the present study we considered three HF methods (BU, TD, and COM), as described in Section \ref{sec:HFmodels}. The reasoning is two-fold. First, classification methods tend to perform better when the number of classes is limited and, as a result, the key differences between the classes are easier to identify \cite{hastie1998classification}. Second, we believe that the selected HF methods are diverse enough, each one focusing on different levels of the hierarchy and adopting a significantly different approach for reconciling the base forecasts. Although we could have considered more HF methods of those proposed in the literature, they are mostly variants of the examined three methods (especially the COM method) and are therefore sufficiently covered.

We should also highlight that the rolling origin evaluation of the off-line phase can be adjusted to any desirable set-up that might be more suitable to the user. For example, if computational cost is not an issue, instead of updating the forecast origin by $h$ periods at a time, a step of one period could be considered to further increase the size of the set used for training the classification method and facilitate learning. The main motivation for considering an $h$-step-ahead update is that this practice suits the way the retail firms operate when forecasting their sales and making their plans, creating also a rich set on which the ML classification method can be effectively trained, without exaggerating in terms of computational cost. Finally, although we chose to train the classification method so that the average accuracy of CHF is minimized across the entire hierarchy, this objective can be easily adjusted in order for CHF to provide more accurate forecasts for a specific level of interest. This choice depends on the decision-makers and can vary based on their focus and objectives. However, we do believe that our choice to optimize forecasting accuracy across the entire hierarchy, weighting equally all hierarchical level, is realistic when dealing with demand forecasting and supply chain management given that in such settings each level supports very different, yet equally important decisions. A similar weighting scheme was adopted in the latest M competition, M5\footnote{https://mofc.unic.ac.cy/m5-competition/}, whose objective was to produce the most accurate point forecasts for 42,840 time series that represent the hierarchical unit sales of ten Walmart stores.

\section{Data and experimental setup}\label{sec:setupdata}

\subsection{Data}\label{sec:dataset}

Although HF is relevant in many applications, such as energy \citep{Spiliotis2020-hj} and tourism \citep{KOURENTZES2019393}, it is most commonly found in the retail industry where SKU demand can be grouped based on location and product-related information. Therefore, the dataset used in this study for empirically evaluating the accuracy of the proposed HF approach involves the sales and prices recorded for 55 hierarchies, corresponding to 55 fast-moving consumer goods (FMCG) of a food manufacturing company sold in various locations in Australia. Although the exact labels of the products are unknown to us, the products include breakfast cereals, long-life milk, and other breakfast products.

\begin{figure}
	\caption{Hierarchical structure of the time series included in the examined dataset.}
	\includegraphics[scale=0.33]{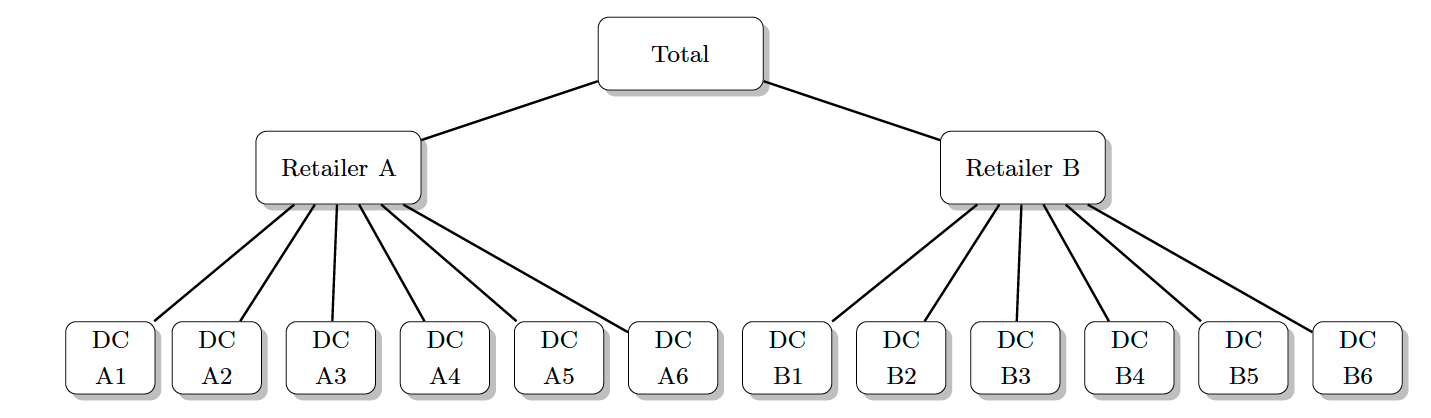}
	\label{htsstructure}
\end{figure}

The hierarchical structure is the same for all 55 products of the dataset and is depicted in Figure \ref{htsstructure}. The number of series at each hierarchical level is provided in Table \ref{table:structure_s}. As seen, each hierarchy consists of three levels, where the top level (level 0) represents total product sales, the middle level (level 1) the product sales recorded for 2 major retailer companies, and the bottom level (level 2) the way the product sales are disaggregated across 12 distribution centers (DC), 6 per retailer company, located in different states of Australia. Thus, each hierarchy includes 15 time series, each containing 120 weeks of observations, spanning from September 2016 to December 2018.

\addtocounter{table}{+1}
\begin{table}
	\centering
	\caption{Number of time series per hierarchical level.}
	\begin{tabular}{l c}\toprule
		\textbf{Hierarchical level } & \textbf{Number of series} \\
		\midrule
		Level 0                      & 1 \\
		Level 1                      & 2 \\
		Level 2                      & 12 \\
		\midrule
		Total                        & 15\\\bottomrule
	\end{tabular}\label{table:structure_s}
\end{table}

\begin{figure}
	\caption{Hierarchical sales of an indicative product of the examined dataset. Level 0 represents total product sales, level 1 the product sales recorded for 2 major retailer companies, and level 2 the product sales across 12 distribution centers.}
	\includegraphics[scale=0.40]{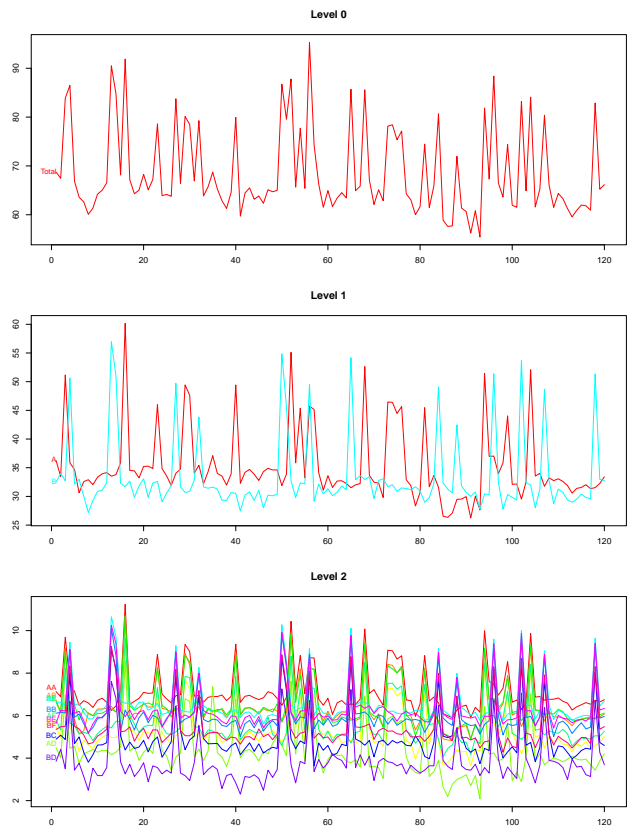}
	\label{sampledata}
\end{figure}

Figure \ref{sampledata} presents the hierarchical time series of an indicative product of the dataset. We observe that sales may experience spikes at different levels of the hierarchy, i.e., different levels of the supply chain. These spikes correspond to promotional periods and their frequency and size vary significantly for different products. Moreover, different nodes in the hierarchy may experience spikes of different extent. As such, the dataset represents a diverse set of demand patterns which are highly affected under the presence of promotions, i.e., price changes.

\subsection{Experimental setup}\label{sec:setup}

Considering that the examined dataset involves products which sales are highly affected by promotions, we produce base forecasts for all the series of the 55 hierarchies using a regression model with ARMA errors (Reg-ARMA), where product prices are used as a regressor variable. Note that Reg-ARIMA can effectively capture sales both during promotional and non-promotional periods as price inherently carries the impact of promotions and, therefore, explains sufficiently the corresponding variations in sales. Reg-ARMA model is implemented using the \textit{forecast} package for R \cite{hyndmanforpack}. 

To evaluate the accuracy of the proposed HF approach both in terms of median and mean approximation \citep{KOLASSA2016788}, we consider two measures, namely the mean absolute scaled error (MASE) and the root mean sum of squared scaled error (RMSSE). The measures are calculated as follows:

\begin{align*}
\text{MASE}  & = \frac{n-1}{h} \frac{ \sum\nolimits_{t=n+1}^{n+h} {|y_{t}-f_{t}|} } {\sum\nolimits_{t=2}^{n} |y_{t}-y_{t-1}|}, \\
\text{RMSSE} & =  \sqrt { \frac{n-1}{h} \frac{ \sum\nolimits_{t=n+1}^{n+h} {(y_{t}-f_{t})^2} } {\sum\nolimits_{t=2}^{n} (y_{t}-y_{t-1})^2} },
\end{align*}

\noindent where $y_t$ and $f_t$ are the observation and the forecast for period $t$, $n$ is the sample size (observations used for training the forecasting model), and $h$ is the forecasting horizon. Note that both measures are scale-independent, thus allowing us to average the results across series.

Once the base forecasts are produced, we use the BU, TD, and COM methods to reconcile them across the three levels of the hierarchy. These methods are used for benchmarking the proposed HF approach as they have been successfully applied in numerous applications and are considered standards in the area of HF \cite{abolghasemi2019machine, hyndman2011optimal, hyndman2016fast}. We also use the CHF algorithm to select which of the three benchmarks is more suitable for forecasting each hierarchy. 

In order to fit and evaluate the CHF algorithm, we split the original dataset into a train and test set. More specifically, we used the first 26 weeks of data to initially train the Reg-ARMA model and the following 58 periods to produce 4-step-ahead base forecasts on a rolling origin basis \citep{TASHMAN2000437}. Each time that the forecast origin was updated, the set used for fitting the Reg-ARMA model was accordingly extended so that the base forecasts produced were appropriately adjusted. Moreover, on each step, the BU, TD, and COM methods were applied to reconcile the base forecasts and identify the most accurate alternative according to MASE. Note that the results were consistent when RMSSE was used as an accuracy measure for training the classifiers, so we report them only for the case of MASE for reasons of brevity. In this respect, a total of 14 accuracy measurements (average accuracy of 4-step-ahead forecasts over 58 weeks) $\times$ 55 hierarchies $\times$ 15 series = 11,550 evaluations were recorded. We then summarised the results across the hierarchy and constructed a dataset of 14 evaluations $\times$ 55 hierarchies = 770 observations that was used for training the XGB classification method. The remaining 36 weeks of data were used as a test set to evaluate the actual accuracy of the proposed approach, again on a rolling origin fashion. Thus, our evaluation is based on a total of 9 accuracy measurements (average accuracy of 4-step-ahead forecasts over 36 weeks) $\times$ 55 hierarchies $\times$ 15 series = 7,425 observations. Note that XGB is trained iteratively each time that we move across the window, with the values of the time series features being accordingly updated.

\section{Empirical results and discussion}\label{sec:hierarchicalresults}

Table \ref{table:results_s} displays the forecasting accuracy (MASE and RMSSE) of the three HF methods considered in this study as benchmarks as we all as classes for training the CHF algorithm in the retail dataset presented in Section \ref{sec:dataset}. The accuracy is reported both per hierarchical level and on average (arithmetic mean of the three levels), while CHF is implemented using the XGB classifier, as described in section \ref{sec:setup}.

\begin{table}
	\centering
	\caption{Forecasting performance of HF methods in terms of MASE and RMSSE.}
	\begin{tabular}{l c c c r} \toprule
		\textbf{HF Method} & \textbf{Level 0} & \textbf{Level 1} & \textbf{Level 2} & \textbf{Average} \\
		\midrule
		\multicolumn{5}{c}{MASE} \\
		\midrule
		BU            &0.502&0.854&0.856&0.832\\
		TD              &\textbf{0.435}&0.987&0.907&0.885\\
		COM       &0.455&0.844 &0.844&0.817\\
		CHF          &0.466&\textbf{0.820}&\textbf{0.828}&\textbf{0.802}\\
		\midrule
		\multicolumn{5}{c}{RMSSE} \\
		\midrule
		BU              &0.583 & 1.049&1.044&1.013\\
		TD              &\textbf{0.507}&1.180&1.100&1.071\\
		COM        &0.531&1.039 &1.035&1.001\\
		CHF           &0.534&\textbf{1.014} &\textbf{1.006} &\textbf{0.975}\\
		\bottomrule
	\end{tabular}\label{table:results_s}
\end{table}

The results indicate that, on average, CHF is the best HF method according to both accuracy measures used. More specifically, CHF provides about 4\%, 9\%, and 2\% more accurate forecasts than BU, TD, and COM, respectively, indicating that the XGB method has effectively managed to classify the hierarchies based on the feautures that their series display. 

The improvements are similar if not better for the middle and bottom levels of the hierarchy. However, CHF fails to outperform TD and COM for the top hierarchical level, being about 6\% and 2\% less accurate, despite being still 8\% better than the BU method. This finding confirms our initial claim that, depending on the hierarchical level of interest, different HF methods may be more suitable. In this study, we focused on the average accuracy of the hierarchical levels and optimised CHF with such an objective. However, as described in section \ref{sec:conditionalhier}, different objectives could be considered in order to explicitly optimise the top, middle or bottom level of the hierarchy.

\begin{figure}
	\centering
	\caption{The ratio of forecasting accuracy of the examined HF methods across all 55 hierarchies included in the dataset. The results are reported for each accuracy measure and hierarchical levels}
	\includegraphics[scale=0.6]{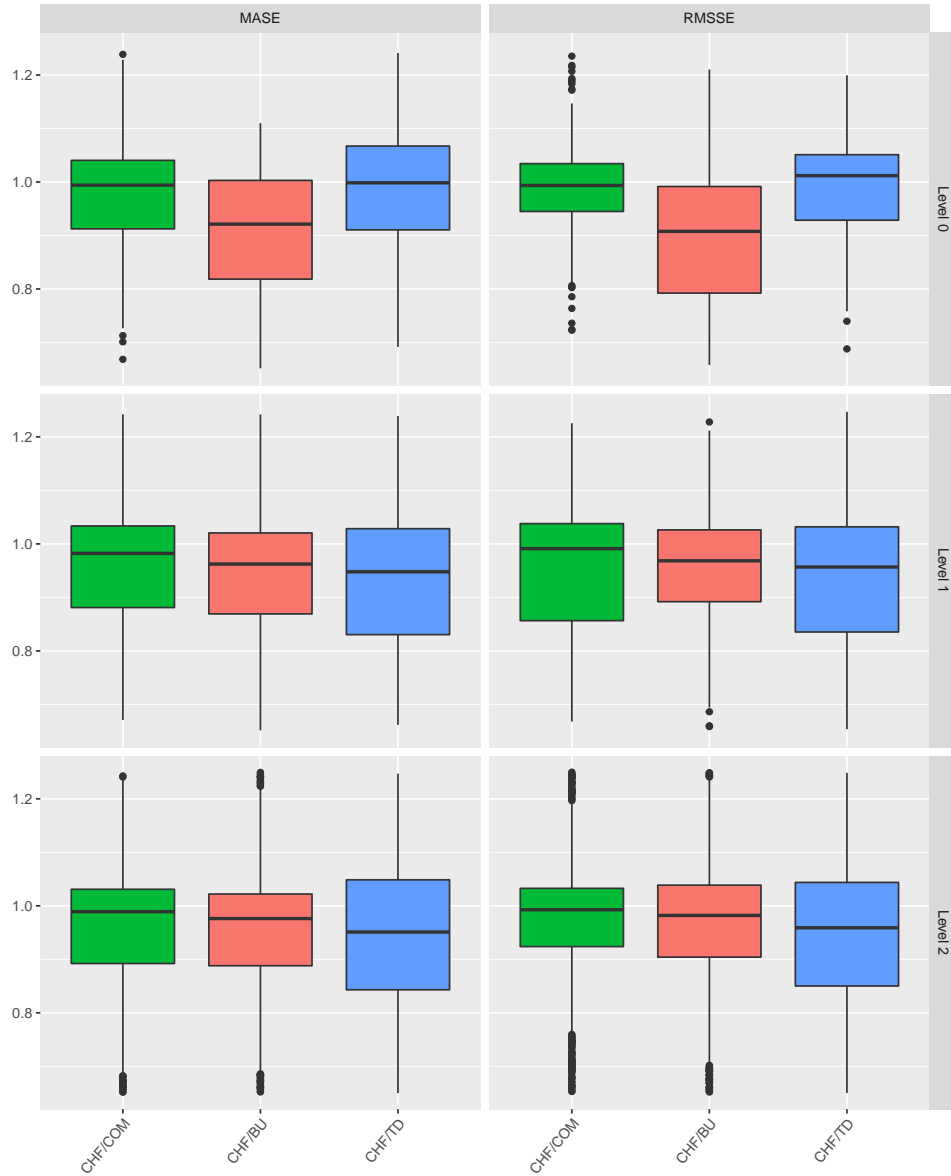}
	\label{fig:ACCAll}
\end{figure}

In order to better evaluate the performance of the proposed approach, we proceed by investigating the distribution of the error ratios reported between CHF and the three benchmark methods examined in our study across all the 55 hierarchies of the dataset. The results, presented per hierarchical level and accuracy measure, are visualised in Figure \ref{fig:ACCAll} where box-plots are used to display the minimum, 1\textsuperscript{st} quantile, median, 3\textsuperscript{rd} quantile, and maximum values of the ratios, as well as any possible outliers. Values lower than unity indicate that CHF provides more accurate forecasts and vice versa. As observed, in most of the cases, CHF outperforms the rest of the HF methods, having a median ratio value lower than unity. The only exception is when forecasting level 0 and using RMSSE for measuring forecasting accuracy, where the TD method tends to provide superior forecasts. Thus, we conclude that CHF does not only provide the most accurate forecasts on average across all the 55 hierarchies, but also the most accurate forecasts across the individual ones.

To validate this finding, we also examine the significance of the differences reported between the various HF methods using the multiple comparisons with the best (MCB) test, as proposed by \citep{koning2005m3}. According to MCB, the methods are first ranked based on the accuracy they display for each series of the hierarchy and then their average ranks are compared considering a critical difference, $r_{\alpha, K, N}$, as follows:

\begin{equation}
r_{\alpha, K, N}= q_a \times \sqrt{\frac{k*(k+1)}{12N}},
\end{equation}\label{MCB_eq}

\noindent where $N$ is the number of the time series, $K$ is the number of the examined HF methods, and $q_a$ is the quantile of the confidence interval. In our case, where $\alpha$ is set equal to 0.05 (95\% confidence), $q_a$ takes a value of 3.219. Accordingly, $K$ is equal to 6 and $N$ is equal to 7,425.

\begin{figure}
	\centering
	\caption{MCB test conducted on the HF methods examined in this study. MASE and RMSSE are used for computing the ranks and a 95\% confidence level is considered.}
	\includegraphics[scale=0.5]{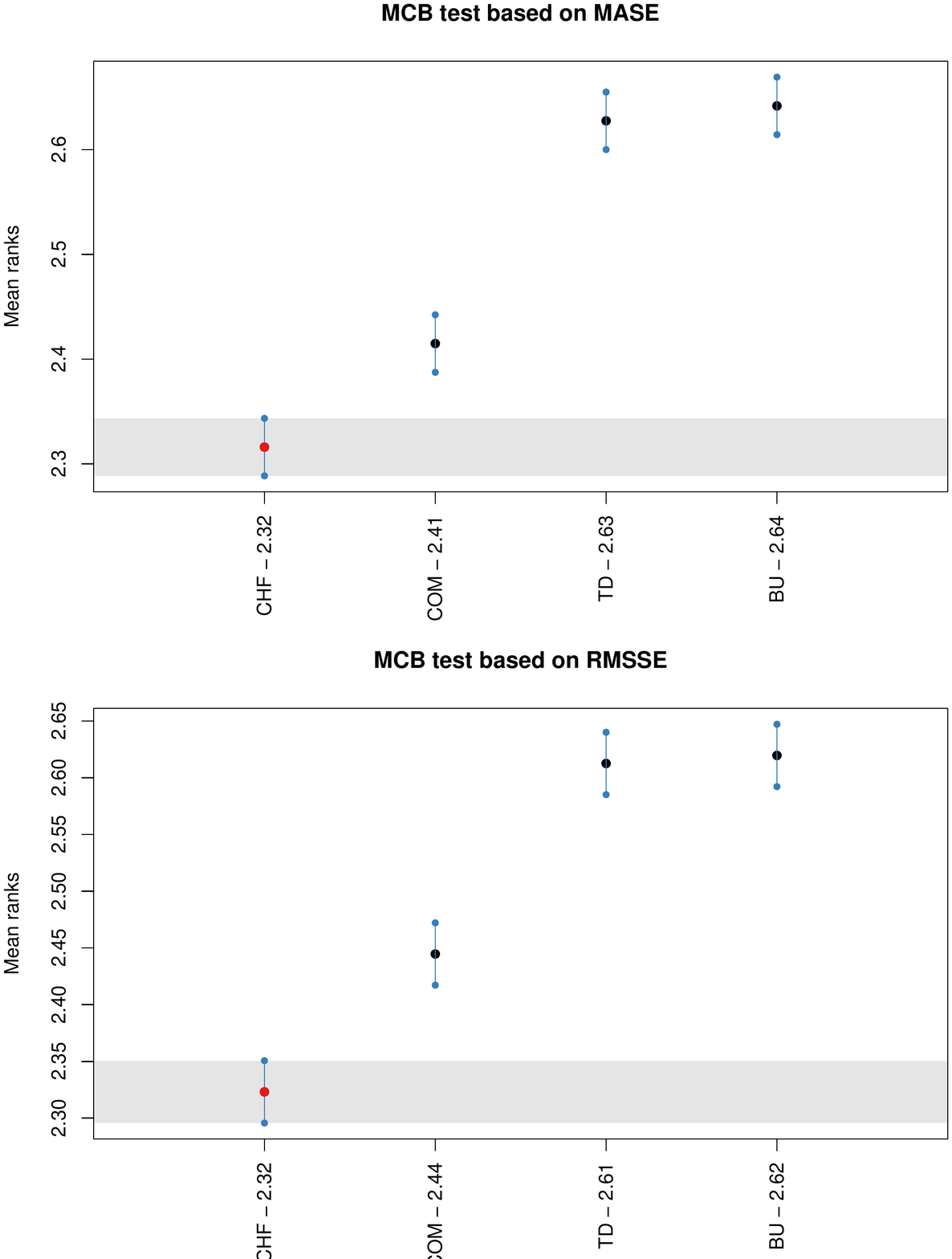}
	\label{fig:MCB}
\end{figure}

The results of the MCB test are presented in Figure \ref{fig:MCB}. If the intervals of two methods do not overlap, this indicates a statistically different performance. Thus, methods that do not overlap with the grey zone of Figure \ref{fig:MCB} are significantly worse than the best, and vice versa. As seen, our results indicate that CHF provides significantly better forecasts than the rest of the HF methods, both in terms of MASE and RMSSE. Moreover, we find that CHF is followed by COM and then by TD and BU. Interestingly, the performance of TD is not significantly different than the one of BU according to RMSSE or MASE. In this regard, we conclude that CHF performs better than the state-of-the-art HF methods found in the literature, being also significantly more accurate than the standards used for HF, such as BU and TD. As such, CHF can be effectively used for selecting the most appropriate HF method from a set of alternatives and improving the overall forecasting accuracy of various hierarchies.

We also investigate the performance of the XGB classifier in terms of precision, recall, and $\text{F}_1$ score. The precision metric measures the number of correct predictions in the total number of predictions made for each class. Recall (also known as sensitivity) informs us about the number of times that the classifier has successfully chosen the best HF method for each class. Finally, the $\text{F}_1$ score is the harmonic mean of precision and recall, computed by $F_1= 2\frac{p*r}{p+r}$, and is used to combine the information provided by the other two metrics \citep{sokolova2009systematic}. 

Before presenting the results, we should first note that the TD, BU, and COM methods have been identified as the most accurate HF method in 1,860, 2,745, and 2,820 cases, respectively. This indicates that the dataset used for training the classification method was equally populated, displaying a uniform probability distribution. Having a balanced training sample is important for our experiment since it facilitates the training of the ML algorithm (enough observations from each class are available and biases can be effectively mitigated) and provides more opportunities for accuracy improvements (if no HF method is dominant, selecting between different HF methods becomes promising).

The performance of the XGB classification method is presented in Table \ref{tab:class_acc}. Observe that, according to the precision metric, XGB manages to select quite accurately the BU and COM methods but finds it difficult to make appropriate selections when the TD method is the most preferable one. This indicates that, although XGB identifies the conditions under which the BU or COM methods are more preferable than the TD method, the opposite is not true.

\begin{table}[!h]
	\centering
	\caption{Performance of the XGB classification method.}
	\begin{tabular}{l c cc r} \toprule
		\textbf{	ML classifier}&\textbf{Class} & \textbf{Precision} & \textbf{Recall} & \boldmath{$F_1$} \\\midrule
		
		
		&BU            & 0.55 & 0.46 & 0.50\\
		XGB&TD       & 0.22 & 0.32 & 0.26 \\
		&COM      &0.42 & 0.38 & 0.40  \\
		
		\bottomrule
	\end{tabular}\label{tab:class_acc}
\end{table}

As a final step in our analysis we investigate the significance of the time series features used by the XGB classifier, i.e., the number of times that each feature was considered by the method for making a prediction. Figure \ref{fig:topfeatures} shows the top-5 selected features of the classifier. As seen, the non-linearity at level 2 is the most frequently used feature and, therefore, the most critical variable for selecting a HF method. This is because non-linearity is one of the strongest features in our dataset, with promotions frequently affecting the sales of the products strongly and changing the volatility of their demand not only over promotional periods but over a number of periods \cite{abolghasemi2020demand}. Thus, the non-linearity at level 1 and level 0 are also among the top-selected features. Maximum variance shift at level 0 is another feature that is frequently selected. This may attribute to the sudden changes and spikes in sales data set caused by promotion. Finally, \emph{e-acf-a.L1}, which contains the first autocorrelation coefficient of the error terms of series at level 1, is also among top selected features. One explanation is because sales are affected by promotions and have a high level of entropy.  
\begin{figure}
	\centering
	\caption{The five features more frequently used by the XGB classification method.}
	\includegraphics[scale=0.45]{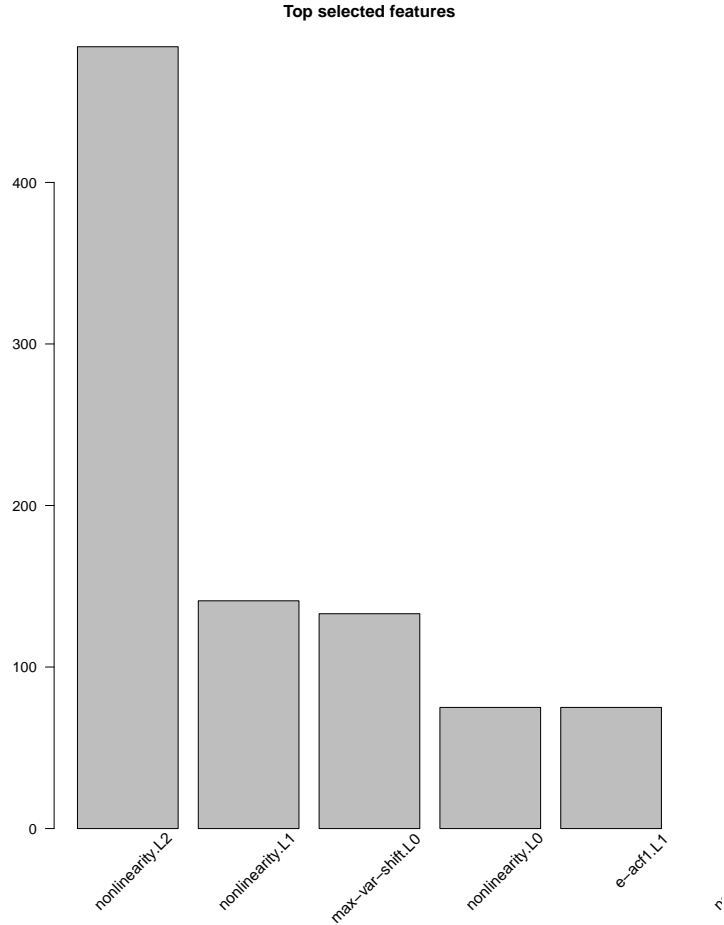}
\end{figure} \label{fig:topfeatures}


Note that when conducting this experiment we considered another training set-up where for the classifiers, apart from time series features, we also provided information about the correlation of the series, both across levels and within each level separately. The results were similar to those reported in Table \ref{table:results_s} and therefore we decided to exclude those features from our models for reasons of simplicity. However, we do believe that features related with the structure and the relationships of the series could help further improving the accuracy of CHF and leave this for future research.

Another extension of the proposed approach that could be also examined in a future study, is that it focuses on selecting the most appropriate hierarchical forecasting method per hierarchy. However, numerous empirical studies have shown that combining forecasts from multiple forecasting methods can improve forecasting accuracy \citep{MAKRIDAKIS202015}. Thus, replacing classifiers with other methods that would combine various HF methods using appropriate weights becomes a promising alternative to CHF. Simple, equal-weighted combinations of standard HF methods have already been proven useful under some settings \citep{Abouarghoub2018-wz}, while feature-based forecast model averaging has demonstrated its potential to generate robust and accurate forecasts \citep{MONTEROMANSO202086}.

\section{Conclusion}\label{sec:hierarchicalconclusion}

This paper introduced conditional hierarchical forecasting, a dynamic approach for effectively selecting the most accurate method for reconciling incoherent hierarchical forecasts. Inspired by the work done in the area of forecasting model selection and the advances reported in the field of Machine Learning, the proposed approach computes various features for the time series of the examined hierarchy and relates their values to the forecasting accuracy achieved by different hierarchical forecasting methods, like bottom-up, top-down, and combination methods, using an appropriate classification method. Based on the lessons learned, and depending on the characteristics of time series in the hierarchy, the most suitable hierarchical forecasting method can be chosen and used to enhance overall forecasting performance. 

We exploited various time series features at different levels of the hierarchy that represent their behavior and trained an extreme gradient boosting classification model to choose the most appropriate type of hierarchical forecasting method for a hierarchical time series with the selected features. The accuracy of the proposed approach was evaluated using a large dataset coming from the retail industry and compared to that of three popular hierarchical forecasting methods. Our results indicate that conditional hierarchical forecasting can produce significantly more accurate forecasts than the benchmarks considered, especially at lower hierarchical levels. Thus, we suggest that, when dealing with hierarchical forecasting applications, selection should be expanded from forecasting model to reconciliation methods as well. 
 
\printbibliography
\end{document}